\documentclass{INTERSPEECH2023}

\usepackage{amsmath,graphicx}
\usepackage[multiple]{footmisc}
\usepackage{multirow}
\usepackage{multicol}
\usepackage{graphicx}
\usepackage{amssymb}

\interspeechcameraready

\title{Unsupervised Out-of-Distribution Dialect Detection with Mahalanobis Distance}
\name{Sourya Dipta Das, Yash Vadi, Abhishek Unnam, Kuldeep Yadav}
\address{
  SHL Labs, India
}
\email{Sourya.Das@shl.com, yash.vadi@yahoo.com, Abhishek.Unnam@shl.com, Kuldeep.Yadav@shl.com}

\begin{document}

\maketitle
 
\begin{abstract}
Dialect classification is used in a variety of applications, such as machine translation and speech recognition, to improve the overall performance of the system. In a real-world scenario, a deployed dialect classification model can encounter anomalous inputs that differ from the training data distribution, also called out-of-distribution (OOD) samples. Those OOD samples can lead to unexpected outputs, as dialects of those samples are unseen during model training. Out-of-distribution detection is a new research area that has received little attention in the context of dialect classification. Towards this, we proposed a simple yet effective unsupervised Mahalanobis distance feature-based method to detect out-of-distribution samples. We utilize the latent embeddings from all intermediate layers of a wav2vec 2.0 transformer-based dialect classifier model for multi-task learning. Our proposed approach outperforms other state-of-the-art OOD detection methods significantly.
\end{abstract}

\section{Introduction}
Dialect identification~\cite{torres2004dialect} has received a lot of interest in the speech processing community in recent decades. Dialect identification plays an important role in speech processing systems such as automated speech recognition (ASR)~\cite{das2021multi}, multilingual translation systems, targeted advertising, and biometric authentication since it helps to target certain dialects. In recent years, numerous approaches~\cite{zhang2018language, kong2021dynamic} have been proposed with great success, for dialect identification. 
However, almost all of these latest state-of-the-art approaches only address closed-set dialect identification, where the set of dialects to predict is fixed. For every input speech data, the predicted dialect within that set is returned. But in a real-life scenario, deployed applications rarely receive regulated inputs from known dialects and are vulnerable to an ever-changing set of unlabeled user inputs with unknown dialects. To solve this issue, a system should have the option to 'reject' that prediction and identify when the input speech does not match any known dialects well. It can also be used to identify and learn new dialects for the system. This task is defined as out-of-distribution(OOD) detection for dialect identification and is essential to the design of trustworthy AI applications in real-world use cases~\cite{hendrycks2016baseline}. The out-of-distribution (OOD) detection problem, in general, has received a great deal of attention in the literature, with cutting-edge algorithms~\cite{lee2018simple,hendrycks2018deep,hendrycks2020pretrained} being supervised in the sense that they require fine-tuning on OOD data to accomplish high performance in OOD detection. Nevertheless, supervised OOD detection algorithms have the problem of requiring expensive training on OOD data, curating the OOD dataset with diverse samples to make it more distinguishable from the in-distribution data, and additional model hyperparameter tuning.

In this paper, we present a joint framework for both dialect classification that automatically classifies known dialects from input speech and out-of-distribution detection which also detects input audio that does not belong to any of the dialects used to train the model. We used a pre-trained wav2vec 2.0 model and fine-tuned it on the known dialects to adapt the feature embedding. This model is used for dialect identification tasks for input data with known dialects. Further, we retrieved features from several intermediate transformer layers to capture rich micro and macro phonetic feature information that may outperform the last layer of the fine-tuned transformer model. Later, we used those latent representations to estimate the mean, and covariance matrix for each layer by using close-set training  data only. During inference, we used previously calculated layer-wise mean and covariance matrices to compute the Mahalanobis distance score for each layer, which was then used as a feature vector. This allows us to use classical outlier rejection methods like KNN~\cite{10.1145/335191.335437} to enumerate outlier scores for the OOD detection task. We have evaluated our proposed method by comparing it with state-of-the-art out-of-distribution detection methods~\cite{liang2018enhancing, shu-etal-2017-doc, bendale2016towards, lee2018simple, liu2020simple, ren2021simple}.
Our contributions can be summarized as follows: 
\begin{itemize}
    \item We present the first joint dialect identification framework with unsupervised OOD dialect detection, which is a plug-and-play technique to identify known dialects or reject input speech samples of unknown dialects in a single forward pass.
    \item We propose Mahalanobis's distance-based feature to be used by a KNN-based~\cite{10.1145/335191.335437} outlier classification model for the OOD dialect identification task. This can be used without modifying the backbone architecture, unlike previous approaches~\cite{bendale2016towards, liu2020simple} that required us to modify the model architecture.
    \item We evaluated the performance of our solution on two different language datasets i.e. English and Spanish to see how well it rejects unknown classes while maintaining its performance on close-set dialect classification. 
\end{itemize}
The proposed approach outperforms the state-of-the-art method and achieves an overall AUROC of 96\% for the \textit{English Dialect Dataset} and 80\% AUROC for the \textit{Spanish Dialect Dataset}. 

\section{Related Work}
There has been existing research work on speech dialect identification, but no prior work on out-of-distribution dialect detection. Torres \textit{et al.}~\cite{torres2004dialect} had done earlier work on dialect identification where they used the Gaussian mixture model (GMM) with shifted delta cepstral features (SDC). Zhang \textit{et al.}~\cite{zhang2018language} proposed an unsupervised bottleneck feature extraction approach for phonetic label estimation. They also used an alternate autoencoder and an adversarial autoencoder in the alternative phase of the speech feature extraction process. Kong \textit{et al.}~\cite{kong2021dynamic} proposes a new model architecture that consists of dynamic kernel convolution, local multi-scale learning, and global multi-scale pooling layers. These proposed custom layers are used to capture features in both short-term and long-term contexts, multiscale granular features from wide receptive fields, and aggregated features from different bottleneck layers, respectively. Hamalainen \textit{et al.}~\cite{hamalainen2021finnish} worked on a Finnish dialect identification system that used speech recording and transcription data. Similarly, Imaizumi \textit{et al.}~\cite{imaizumi2022end}, Ma \textit{et al.}~\cite{ma2006chinese} and Lin \textit{et al.}~\cite{lin2020transformer} also worked on Japanese, Chinese and Arabic dialects identification techniques, respectively.

Recently, there have been several studies that have explored the out-of-distribution problem in various domains like vision, text, etc. Liang \textit{et al.}~\cite{liang2018enhancing} have shown that thresholding onto the softmax output of the predicted class provides a good proxy score for detecting Out of Distribution (OOD) data. Shu \textit{et al.}~\cite{shu-etal-2017-doc} suggested another approach called DOC (Deep Open Classification). In contrast to conventional classifiers, DOC constructs a multi-class classifier with a 1-vs-rest final layer of sigmoids instead of a softmax to minimize the risk associated with open spaces. By reducing the decision bounds of sigmoid functions with Gaussian fitting, it significantly lowers the open space risk for rejection. 
Bendale \textit{et al.}~\cite{bendale2016towards} then presented a new neural network layer, OpenMax, which estimates the likelihood that an input belongs to an unknown class. They estimate the unknown class rejection probability value by adapting the extreme-value Meta-Recognition-inspired distance normalization process to the activation patterns in the penultimate layer of the network.
Lee \textit{et al}.~\cite{lee2018simple} proposed building a Gaussian model from features extracted from the hidden layer and calculating the distance from this multivariate distribution (Mahalanobis Distance) and used this distance for OOD detection. Ren \textit{et al.}~\cite{ren2021simple} modified the Mahalanobis distance by subtracting the distance calculated from the entire training distribution to make it suitable for detecting near OOD samples. Liu \textit{et al.}~\cite{liu2020simple} has developed a robust uncertainty-based methodology that delivers an uncertainty score to each prediction and may be used to discover outliers.
Here, we compared recent state-of-the-art machine learning and deep learning-based OOD detection techniques, as stated above, in an audio dialect setting.

\section{Problem Statement}
The problem is formulated as a variant of conventional multi-class classification which is also referred to as a close-set classification problem. 
Given dialect classification training data $D_{train} = \{(x_1, y_1), (x_2, y_2) \dots (x_N, y_N)\}$ where $N$ is the total number of samples in training data, $x_i$ is the input audio sample, $y_i \in L_{knw} = \{1, …,M\}$ is corresponding target dialect label with $M$ number of dialect classes, contains samples from a fixed set of known dialect classes. During inference, the test set contains samples from both the set of known dialect classes during training and additional unknown dialect classes, i.e., $D_{test} = \{(x'_1, y'_1), (x'_2, y'_2) \dots (x'_n, y'_n)\}$ where $y'_i \in (L_{knw} \cup L_{unk})$ and $L_{unk}$ includes classes that are not observed during training. 

In this paper, we use a more realistic scenario in which we have no prior knowledge of what out-of-distribution inputs look like. It is not possible to train a separate supervised classifier directly in this scenario. So, our task is to train a classifier $F_D(x)$ with $D_{train}$ training data, that correctly predicts the dialect class from a set of known dialect classes, i.e. $F_D(x) = [d_1,d_2,\dots,d_M]$ where $d_m$ is prediction score of $m$-th known dialect class and accurately detects audio samples with unknown dialect class by classifying those audio samples as a rejected class, which is denoted as the M + 1 class. 
\begin{equation}
\hat{y_p}= \begin{cases}\operatorname{argmax}_{m} F_D(x) & \text { if } G_D(x) \leq \delta \\ M+1 & \text { if } G_D(x) > \delta\end{cases}
\end{equation}
Here, $\hat{y_p}$  is the predicted class and $G_D(x)$ is a class rejection score function that determines if the input corresponds to the unknown dialect class or rejected class and $\delta$ is a threshold value. Here, the OOD dialect detection problem is as simple as constructing a class rejection score function, $G_D(x)$ that assigns lower scores to inputs with known dialect class than to out-of-distribution inputs.

\section{Methodology}
\subsection{Model Architecture and Fine-Tuning}
Wav2Vec 2.0 speech model~\cite{baevski2020wav2vec} is pre-trained on unlabeled speech data using self-supervised learning for learning high-quality representations of speech. It shows promising results when transferred to other tasks~\cite{fan2020exploring, li2021accent} like speech classification, speech recognition, speech frame classification, etc. Therefore, we have used a pre-trained wav2vec 2.0 model and have fine-tuned it on the $D_{train}$ dataset for closed set known dialect classification on $M$ classes for learning the feature embeddings on the dataset. After fine-tuning, we obtain a wav2vec 2.0 architecture-based dialect identification model, $F_D$ with $K$ transformer layers.
\subsection{Class Rejection Score Estimation Method}
\begin{figure}[h]
  \centering
  \includegraphics[width=\linewidth]{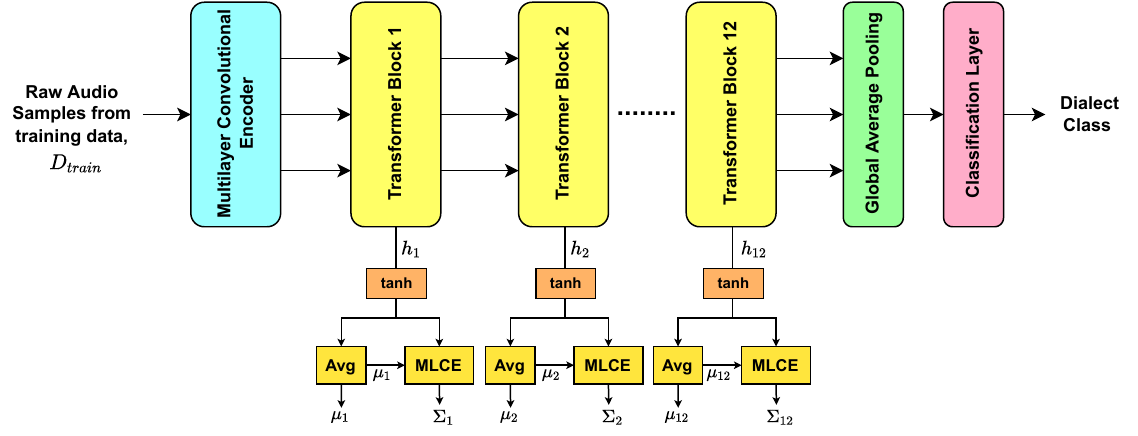}
  \caption{Illustration of Layer Feature Embedding Mean($\mu_k$), Covariance Matrix ($\Sigma_k$) Estimation for $k$-th transformer layer from the feature embeddings,$F^{k}_D(x)$ of training data, $D_{train}$. Here, \textit{Avg} is component-wise vector average operation and \textit{MLCE} is Maximum Likelihood Covariance Estimator.}
  \label{fig:MuSigma_wav2vec2}
\end{figure}

\begin{figure}[t]
  \centering
  \includegraphics[width=\linewidth]{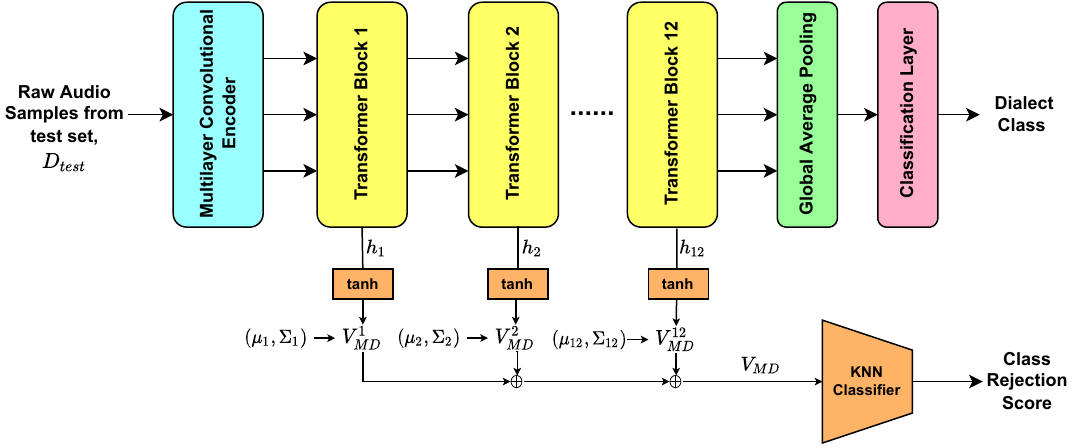}
  \caption{OpenSet wav2vec 2.0 Dialect Classifier Architecture. Here, $\oplus$ is concatenation operator and $V_{MD}(x) = [V_{MD}^1(x)\oplus V_{MD}^2(x)\oplus\dots\oplus V_{MD}^K(x)]$, is Mahalanobis Distance Feature Vector.}
  \label{fig:OpenSet_wav2vec2}
\end{figure}

 We denote $F^{k}_D(x) \in \mathbb{R}_d$ as the $d$-dimensional feature embeddings corresponding to the $k$-th transformer layer for input $x$ where $k \in [1,2, \dots,K]$.
 we further passed those intermediate feature embeddings through a hyperbolic tangent function, $tanh(.)$ to transform the features into the same restricted semantic space, i.e., $h^{t}_k(x) = tanh(F^{k}_D(x))$. 
 We used this function to limit the value of each embedding vector element to between +1 and -1, similar to feature value scaling.
From a previous study~\cite{shah2021all}, different transformer layers of the wav2vec 2.0 model capture distinct semantic properties from the input speech. Thus, we use latent representations from all transformer layers of fine-tuned wav2vec 2.0 dialect classifier model by concatenating feature embeddings
from all transformer layers, i.e., $\phi_h(x) = [h^{t}_1, h^{t}_2,\dots, h^{t}_K]^T \in \mathbb{R}_{d.K}$. 
From previous work~\cite{lee2018simple}, we use Mahalanobis distance to calculate the distance between test audio samples and training data distribution, $D_{train}$ for detecting unknown classes. Here, we defined the Mahalanobis distance score by using a simple and computationally efficient approximation method in a prior work~\cite{xu2021unsupervised}. We achieved that by decomposing the feature space into several subspaces and solving a low-dimensional constrained convex optimization. We illustrate this estimation process in Figure \ref{fig:MuSigma_wav2vec2}. Thus, we define Mahalanobis distance score, $V_{MD}^k(x_i)$ in the following equation.
\begin{gather*}
\small
    V_{MD}^k(x_i)=\left(h^{t}_k(x_i))-\mu_k\right)^T\Sigma_k^{-1}\left(h^{t}_k(x_i))-\mu_k\right) \\
    \mu_k=\frac{1}{n} \sum_{i=1}^{n}\left[h^{t}_k(x_i))\right] \\
    \Sigma_{k} =\frac{1}{(n-1)w_k} \sum_{i=1}^{n}\left(h^{t}_k(x_i)-\mu_k\right)\left(h^{t}_k(x_i)-\mu_k\right)^{T}
\end{gather*}
where $\mu_k, \Sigma_k$ are mean and covariance for $k$-th transformer layer from the feature embeddings of training data, $D_{train}$ respectively, $w_k$ is a layer-dependent constant from that optimization process for $k$-th transformer layer and the square root of $V_{MD}^k(x_i)$ is the Mahalanobis distance of the transformer layer embedding of data $x_i$ from the $k$-th layer. We enumerate the value of $w_k$ during that optimization process to extract relevant hidden state features from transformer layer embeddings. we further define Mahalanobis distance feature vector, $V_{MD}(x) = [V_{MD}^1(x)\oplus V_{MD}^2(x)\oplus\dots\oplus V_{MD}^K(x)]$ by concatenating Mahalanobis distance scores, $V_{MD}^k(x)$ for all transformer layers. Then, we train a KNN~\cite{10.1145/335191.335437} based outlier detection model with Mahalanobis distance feature vectors extracted from training samples to estimate Class Rejection Score, $G_D(x)$ for detection of unknown class with a threshold value, $\delta$. We illustrate the Inference pipeline of the proposed method in Figure \ref{fig:OpenSet_wav2vec2}.

\section{Experimentation and Results}
We implemented our dialect classifier model using PyTorch on
top of the Wav2Vec 2.0 model from Hugging Face transformer library~\footnote{https://huggingface.co}. 
We have used two different pre-trained wav2vec 2.0 model~\footnote{https://huggingface.co/facebook/wav2vec2-base}\footnote{https://huggingface.co/facebook/wav2vec2-base-10k-voxpopuli-ft-es} for English and Spanish dialect datasets. Furthermore, we have used all 12 transformer layers of our Wav2Vec 2.0 base model for defining the Mahalanobis distance score in all of our experiments. For class rejection score estimation, we have used PyOD \footnote{https://pyod.readthedocs.io/en/latest/} package to implement KNN~\cite{10.1145/335191.335437} with 0.01 outlier fraction value and scikit-learn~\footnote{https://scikit-learn.org/} for implementing Maximum likelihood covariance estimator(MLCE).  We train each dialect classification model with 6 epochs using 1 NVIDIA GTX 1080 GPU (12 GB) with 16 GB RAM. For both training and validation, all experiments use only close-set training data with fixed known classes. 

\subsection{Dataset Details}
We have used two custom-made datasets, \emph{English Dialect Dataset}, and \emph{Spanish Dialect Dataset} for the two most spoken languages, English and Spanish respectively to evaluate our method. We have sampled speech data from  \emph{AccentDB}~\cite{ahamad-anand-bhargava:2020:LREC}, \emph{UK and Ireland English Dialect speech dataset}~\cite{demirsahin2020open} and \emph{Google Nigerian English speech dataset}\footnote{https://openslr.org/70/} for custom \emph{English Dialect Dataset}. We have used whole \emph{Latin American Spanish speech dataset}~\cite{guevara2020crowdsourcing} for our custom \emph{Spanish Dialect Dataset}. During training, we hide a few classes and used those hidden classes as unknown classes in the test set for efficacious open-world evaluation. The details on these custom speech datasets are following. 

\textbf{\textit{English Dialect Dataset }} consists of 11383 audio samples (spoken by 80 speakers) with 4 classes which are `Southern', `Northern', `Welsh', and `Scottish', used as fixed known classes, and 4800 audio samples (spoken by 12 speakers) with 4 classes which are `Indian', `American', `Nigerian', and `Australian' used as unknown class samples in the test set for OOD dialect detection evaluation. Close-set data is a subset of \emph{UK and Ireland English Dialect speech dataset} and outlier or out-of-distribution  samples are  from both \emph{AccentDB} and \emph{Google Nigerian English speech dataset}. More details on the respective train, validation, and test set are provided in Table \ref{datset_stat}.

\textbf{\textit{Spanish Dialect Dataset}} consists of 17724 audio samples spoken by 79 speakers with known 4 classes: `Argentinian', `Peruvian', `Colombian', and `Chilean', used as fixed known classes, and 3674 audio samples (spoken by 23 speakers) with 2 classes which are `Venezuelan' and `Puerto rico', which are used as unknown class test samples in the test set for OOD dialect detection evaluation. More details on the respective train, validation, and test set are provided in Table \ref{datset_stat}.

\begin{table}
\centering
\scriptsize
\caption{Details of Speech Dialect Classification Datasets for Out of Distribution Detection.}
\label{datset_stat}
\begin{tabular}{|c|c|c|c|} 
\hline
\begin{tabular}[c]{@{}c@{}}\textbf{Dialect }\\\textbf{Dataset}\end{tabular}         & \begin{tabular}[c]{@{}c@{}}\textbf{Dataset }\\\textbf{Split}\end{tabular} & \begin{tabular}[c]{@{}c@{}}\textbf{No of }\\\textbf{Samples}\end{tabular} & \begin{tabular}[c]{@{}c@{}}\textbf{Time Duration}\\\textbf{~(in Hours)}\end{tabular}  \\ 
\hline
\multirow{4}{*}{\textit{English}} & Train-set                                                                 & 9738                                                                      & 17.194                                                                                \\ 
\cline{2-4}
                         & Validation-set                                                            & 550                                                                       & 1                                                                                     \\ 
\cline{2-4}
                         & Test-set (known dialects)                                                              & 5895                                                                      & 2.004                                                                                 \\ 
\cline{2-4}
                         &  Test-set (unknown dialects)                                                             & 4800                                                                      & 5.5433                                                                                \\ 
\hline
\multirow{4}{*}{\textit{Spanish}} & Train-set                                                                 & 13715                                                                     & 21.29                                                                                 \\ 
\cline{2-4}
                         & Validation-set                                                            & 2010                                                                      & 3.233                                                                                 \\ 
\cline{2-4}
                         &  Test-set (known dialects)                                                              & 5973                                                                      & 3.213                                                                                 \\ 
\cline{2-4}
                         &  Test-set (unknown dialects)                                                             & 3974                                                                      & 5.817                                                                                 \\
\hline
\end{tabular}
\end{table}

\subsection{Evaluation Metric}
Known dialect classification performance is measured using precision, recall, and F1 scores. For the detection of OOD dialects or unknown dialects, we are using the evaluation metrics that have been previously used in \cite{ryu-etal-2018-domain, liang2018enhancing, lee2018simple} because this can be considered as an out-of-distribution detection. Specifically, TP, TN, FP, FN, TPR, and FPR represent true positive, true negative, false positive, false negative, true positive rate, and false positive rate respectively. We use the following metrics for OOD dialect detection evaluation:

\textbf{AUROC (Higher is better)}
is the area under the Receiver Operating Characteristic (RoC) Curve. The RoC is plotted TPR against FPR by varying the threshold.

\textbf{AUPR (Higher is better)}
is the area under the curve plotted precision against the recall by varying the threshold value. AUPR(IN) and AUPR(OUT) represent the fixed known classes and the outlier unknown classes as positive classes respectively.

\textbf{EER (Lower is better)}
is the error rate of the classifier when the confidence threshold is set where the FPR (FPR = FP/(FP+TN)) is equal to FNR (FNR = FN/(FN+TP)).
\begin{equation}
\small
\mathrm{EER}=\frac{\mathrm{FP}+\mathrm{FN}}{\mathrm{TP}+\mathrm{TN}+\mathrm{FN}+\mathrm{FP}}
\end{equation}

\subsection{Close-set Performance Results}
Table \ref{in_domain_results} shows the performance of the proposed model on known dialect categorization tasks for both datasets. These results show that the proposed method does not compromise the accuracy of the model in dialect classification tasks for known dialect classes.
\begin{table}[h]
\scriptsize
\caption{Close-set Dialect Classification Performance Results}
\label{in_domain_results}
\centering
\begin{tabular}{|c|c|c|c|}
\hline
\textbf{Dataset} & \textbf{Recall} & \textbf{Precision} & \textbf{F1} \\ \hline
English Dialect & 90.1 & 89.7 & 89.23 \\ \hline
Spanish Dialect & 97.77 & 97.51 & 97.57 \\ \hline
\end{tabular}%
\end{table}
\vspace{-0.3cm}
\subsection{Ablation Study}
Here, we experiment with different outlier detection models to show the effectiveness of our proposed model for class rejection score estimation.
For this study, we have used Mahalanobis distance features to train CBLOF (cluster-based local outlier factor) \cite{10.1016/S0167-8655(03)00003-5}, Isolation Forest \cite{Liu_isolationforest}, KNN \cite{10.1145/335191.335437}, local outlier factor \cite{10.1145/335191.335388}, and one-class SVM \cite{article} models with the same setup as our own model for OOD task. 
Table \ref{English_abal_results} and Table \ref{Spanish_abal_results} show that our proposed method delivers the best outcomes in both datasets. 

\begin{table}[h]
\scriptsize
\centering
\caption{Ablation study results of English Dialect Dataset}
\label{English_abal_results}
\begin{tabular}{|c|c|c|c|c|}
\hline
\textbf{Methods} & \textbf{EER} & \textbf{AUROC} & \textbf{\begin{tabular}[c]{@{}c@{}}AUPR\\ (IN)\end{tabular}} & \textbf{\begin{tabular}[c]{@{}c@{}}AUPR\\ (OUT)\end{tabular}} \\ \hline
CBLOF~\cite{10.1016/S0167-8655(03)00003-5} & 0.1625 & 92.12 & 76.09 & 9778 \\ \hline
\begin{tabular}[c]{@{}c@{}}Isolation\\ Forest~\cite{Liu_isolationforest}\end{tabular} & 0.1351 & 94.14 & 82.93 & 98.43 \\ \hline
LOF~\cite{10.1145/335191.335388} & 0.2146 & 85.55 & 59.52 & 95.36 \\ \hline
OC-SVM~\cite{article} & 0.9762 & 51.19 & 60.25 & 90.89 \\ \hline
Our Method~\cite{10.1145/335191.335437} & \textbf{0.0959} & \textbf{96} & \textbf{86.78} & \textbf{98.81} \\ \hline
\end{tabular}
\end{table}


\begin{table}[h]
\scriptsize
\centering
\caption{Ablation study results of Spanish Dialect Dataset}
\label{Spanish_abal_results}
\begin{tabular}{|c|c|c|c|c|}
\hline
\textbf{Methods} & \textbf{EER} & \textbf{AUROC} & \textbf{\begin{tabular}[c]{@{}c@{}}AUPR\\ (IN)\end{tabular}} & \textbf{\begin{tabular}[c]{@{}c@{}}AUPR\\ (OUT)\end{tabular}} \\ \hline
CBLOF~\cite{10.1016/S0167-8655(03)00003-5} & 0.2836 & 77.8 & 64 & 86.02 \\ \hline
\begin{tabular}[c]{@{}c@{}}Isolation\\ Forest~\cite{Liu_isolationforest}\end{tabular} & 0.2791 & 78.36 & 66.47 & 85.52 \\ \hline
LOF~\cite{10.1145/335191.335388} & 0.4412 & 58.07 & 39.88 & 71.70 \\ \hline
OC-SVM~\cite{article} & 0.6698 & 64.82 & 69.12 & 85.91 \\ \hline
Our Method~\cite{10.1145/335191.335437} & \textbf{0.2726} & \textbf{80.31} & \textbf{71.33} & \textbf{86.76} \\ \hline
\end{tabular}
\end{table}

\subsection{Quantitative Comparison}
We compare our method to other state-of-the-art methods discussed in recent literature and have reported their performance in Table \ref{spanish_comp_results} and Table \ref{English_comp_results} for each dialect dataset.  We use the same Wave2vec2.0 model as the backbone throughout all these comparison methods to validate the comparison setup. From these results, it is very prominent that our proposed method outperforms other methods by considerable margins. 
Since our method makes use of multiple hidden layer embeddings and the KNN classifier model, it outperforms the closest method MD~\cite{lee2018simple} and RMD~\cite{ren2021simple}.
\begin{table}[!h]
\scriptsize
\caption{Quantitative Comparison Results of Spanish Dialect Dataset}
\label{spanish_comp_results}
\centering
\begin{tabular}{|c|c|c|c|c|}
\hline
\textbf{Methods} & \textbf{EER} & \textbf{AUROC} & \textbf{\begin{tabular}[c]{@{}c@{}}AUPR\\ (IN)\end{tabular}} & \textbf{\begin{tabular}[c]{@{}c@{}}AUPR\\ (OUT)\end{tabular}} \\ \hline
Max Thresold~\cite{liang2018enhancing} & 0.4227 & 63.54 & 57.28 & 74.52 \\ \hline
DOC~\cite{shu-etal-2017-doc} & 0.3250 & 55.34 & 66.98 & 79.67 \\ \hline
Openmax~\cite{bendale2016towards} & 0.3585 & 55.81 & 37.84 & 71.56 \\ \hline
MD~\cite{lee2018simple} & 0.3116 & 74.94 & 63.79 & 84.6 \\ \hline
SNGP~\cite{liu2020simple} &  \textbf{0.2496} & 62.39 & 55.31 & 78.65 \\ \hline
RMD~\cite{ren2021simple} & 0.3106 &  75.02 & 63.07 &  84.64 \\ \hline
Our method & 0.2726 & \textbf{80.31} & \textbf{71.33} & \textbf{86.76} \\ \hline
\end{tabular}
\end{table}

\begin{table}[!h]
\scriptsize
\centering
\caption{Quantitative Comparison Results of English Dialect Dataset}
\label{English_comp_results}
\begin{tabular}{|c|c|c|c|c|}
\hline
\textbf{Methods} & \textbf{EER} & \textbf{AUROC} & \textbf{\begin{tabular}[c]{@{}c@{}}AUPR\\ (IN)\end{tabular}} & \textbf{\begin{tabular}[c]{@{}c@{}}AUPR\\ (OUT)\end{tabular}} \\ \hline
Max Thresold~\cite{liang2018enhancing} & 0.1342 & 63.81 & 56.98 & 90.22 \\ \hline
DOC~\cite{shu-etal-2017-doc} & 0.3376 & 53.94 & 52.88 & 94.93 \\ \hline
Openmax~\cite{bendale2016towards} & 0.3468 & 78.34 & 56.51 & 91.54 \\ \hline
MD~\cite{lee2018simple} & 0.3004 & 78.35 & 51.62 & 93.19 \\ \hline
SNGP~\cite{liu2020simple} & 0.1716 & 86.38 & 57.37 & 95.13 \\ \hline
RMD~\cite{ren2021simple} & 0.2876 & 79.97 & 49.63 & 94.17 \\ \hline
Our method &  \textbf{0.0959} & \textbf{96} & \textbf{86.78} & \textbf{98.81} \\ \hline
\end{tabular}
\end{table}

\section{Conclusion}
This paper presents the dialect classification problem in open-world scenarios and proposes a wav2vec 2.0 transformer model-based method to not only recognize dialects known during the training process but also detect unknown dialects as rejected classes at the inference time. We test our approach on two large-scale open-source dialect speech datasets and also present its performance comparison with other methods that are widely used in vision and language processing. Furthermore, our quantitative comparison experiment also indicates that integrating several intermediate layer output vectors to compute Mahalanobis distance-based feature vectors delivers higher performance than other prior Mahalanobis distance-based OOD detection methods~\cite{lee2018simple, ren2021simple}. In future work, we would like to investigate how adversarial training and contrastive learning can be helpful for out-of-distribution dialect classification. 

\bibliographystyle{IEEEtran}
\bibliography{mybib}

\end{document}